\documentclass[runningheads]{llncs}
\usepackage{graphicx}
\usepackage{mystyle}
\usepackage{multirow}
\usepackage{amsmath}
\usepackage{comment}

\usepackage{bm} 
\newcommand{\Bx}{{\bm x}}

\newcommand{\Bd}{{\bm d}}

\newcommand{\Bp}{{\bm p}}

\def\typeG{G}
\def\typeB{BS}

\def\thickhline{\noalign{\hrule height.8pt}}

\newcommand{\gcheck}{{\color{black}\checkmark}}

\begin{document}
%
\title{Test-Time Augmentation \\ for Traveling Salesperson Problem}
%
%

%
\institute{Kyushu University, Fukuoka, Japan}
\author{Ryo Ishiyama \and
Takahiro Shirakawa \and
Seiichi Uchida \and
Shinnosuke Matsuo
}
\authorrunning{R. Ishiyama et al.}
%
%
\maketitle              
\begin{abstract}

We propose Test-Time Augmentation (TTA) as an effective technique for addressing combinatorial optimization problems, including the Traveling Salesperson Problem.
In general, deep learning models possessing the property of invariance, where the output is uniquely determined regardless of the node indices, have been proposed to learn graph structures efficiently.
In contrast, we interpret the permutation of node indices, which exchanges the elements of the distance matrix, as a TTA scheme.
The results demonstrate that our method is capable of obtaining shorter solutions than the latest models.
Furthermore, we show that the probability of finding a solution closer to an exact solution increases depending on the augmentation size.

\end{abstract}


\section{Introduction\label{sec:intro}}
\begin{figure}[t]
    \centering
    \includegraphics[width=0.9\linewidth]{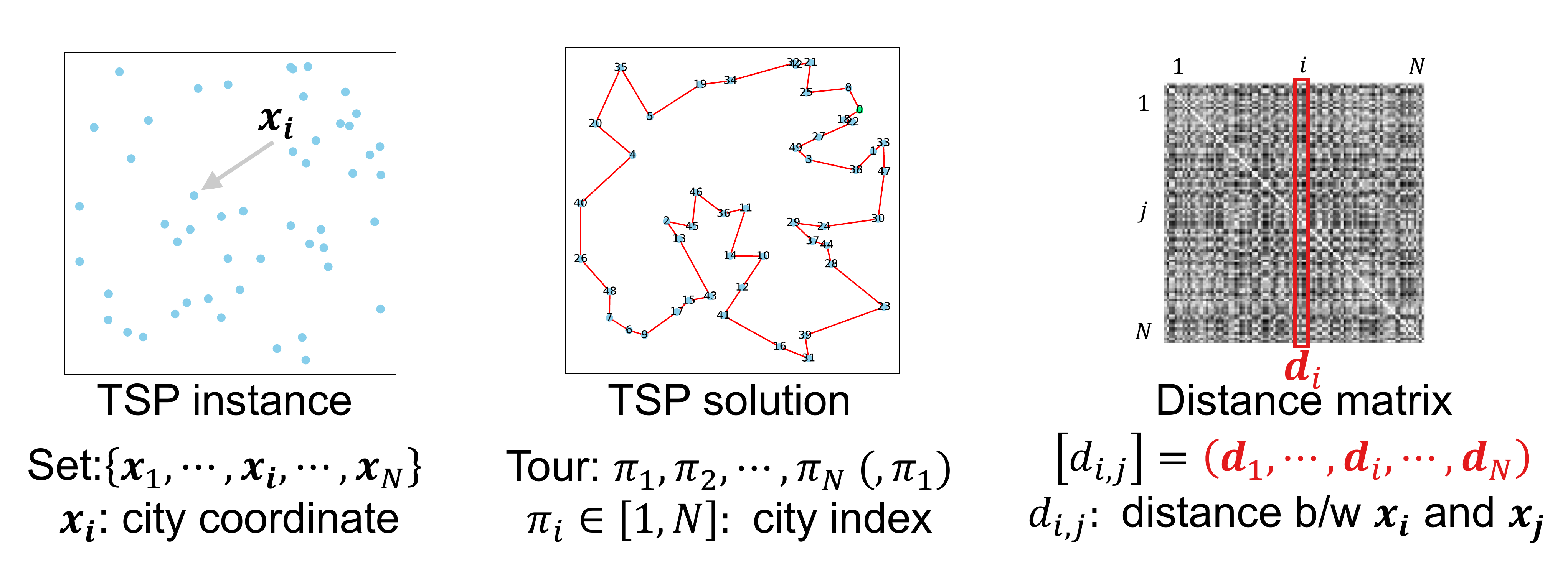}\\[-2mm]
    \caption{An instance and its solution of the (Euclidean) traveling salesperson problem (TSP). Note that its distance matrix depends on the order of cities. \label{fig:tsp_style}}
\end{figure}

The traveling salesperson problem (TSP) is a well-known NP-hard combinatorial optimization problems~\cite{papadimitriou1977euclidean}. TSP is a problem of finding the shortest-length tour\footnote{This paper assumes ``Euclidean'' TSP unless otherwise mentioned.}  (i.e., circuit) that visits each city once. More formally, as shown in Fig.~\ref{fig:tsp_style}, the instance of TSP is a set $\{\Bx_1, \ldots, \Bx_i, \ldots, \Bx_N\}$, where $\Bx_i$ is the coordinate of a city, and its solution is a tour represented as the city index sequence, $\pi_1,\ldots,\pi_i,\ldots,\pi_N$, where $\pi_i\in [1,N]$.
Traditional research focuses on exact solvers~\cite{applegate2006concorde,Held1962ADP} and approximation solvers based on algorithms such as minimum spanning tree, 2-opt~\cite{Christofides1976WorstCaseAO,Helsgaun2000AnEI,Helsgaun2017AnEO,Johnson1990LocalOA,johnson1997traveling,Lin1965ComputerSO}. 
In contrast, modern approximation solvers use machine-learning frameworks
~\cite{Bello2016NeuralCO,bresson2021transformer,Deudon2018LearningHF,hudson2022graph,gcn-tsp,Kaempfer2018LearningTM,Khalil2017LearningCO,kool2018attention,Nazari2018ReinforcementLF,Nowak2017ANO,Vinyals2015}. 
For TSP, it is difficult to use the standard fully-supervised machine-learning framework because the ground-truth of TSP is hardly available due to its computational complexity. Therefore, deep reinforcement learning is often employed because it does not need the ground-truth but only needs the evaluation of the current solution (i.e., tour length) as a reward. 
\par

\begin{figure}[t]
   \centering
    \includegraphics[width=\linewidth]{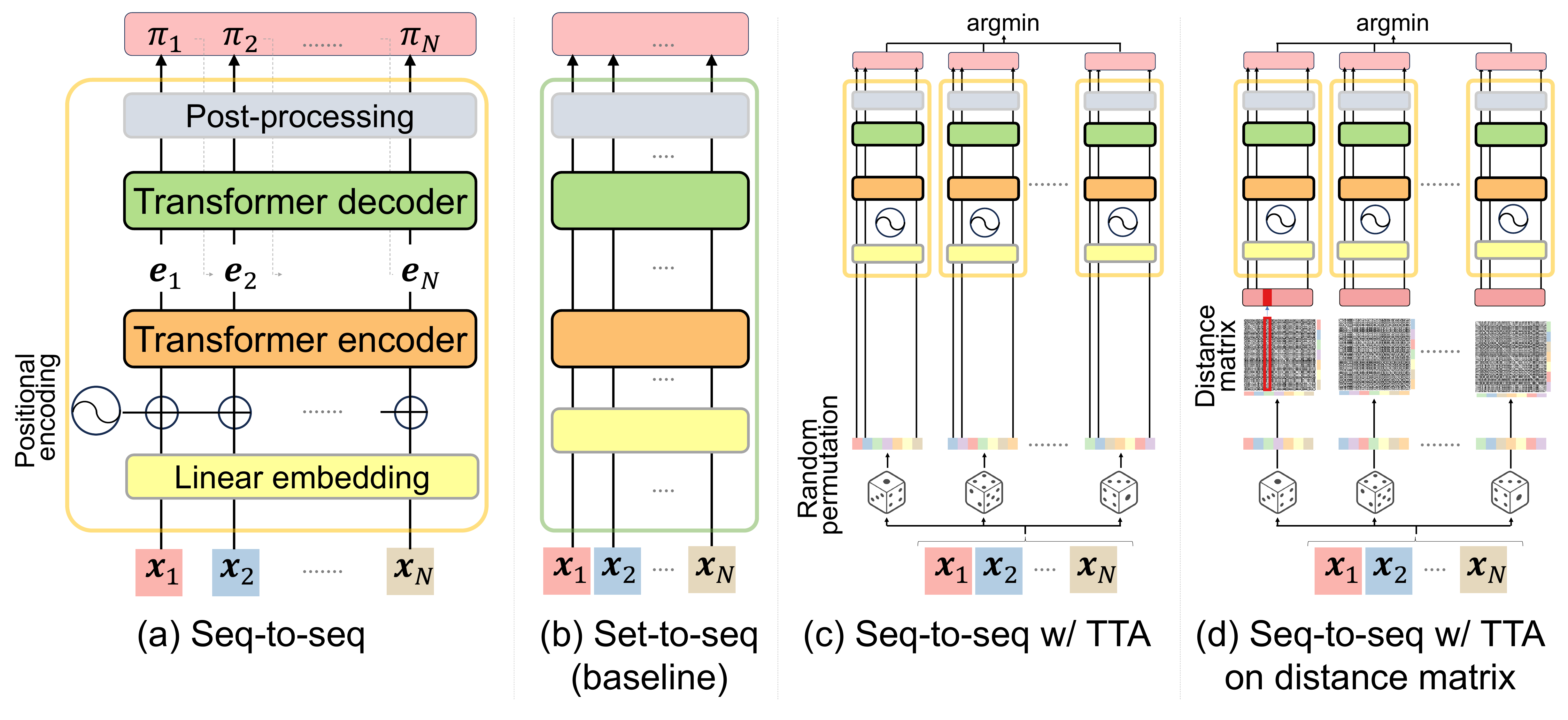}\\[-2mm]
    \caption{Transformer encoder and decoder model. (a)~The standard sequence-to-sequence model with positional encoding. (b) A set-to-sequence model for solving TSP~\cite{kool2018attention,bresson2021transformer}, where positional encoding is removed from (a). 
    (c)~TTA with the $M$ sequence-to-sequence model of (a). (d)~The proposed model, where $M$ distance matrices are used for TTA.
    \label{fig:our_model}}
\end{figure}

The latest and well-known machine learning-based solver is Bresson et al.~\cite{bresson2021transformer}. Roughly speaking, their solver formulates TSP as a sequent-to-sequence conversion task (in fact, it is not exactly true as detailed soon) and solves the task using a Transformer-based solver. Fig.~\ref{fig:our_model}(a) illustrates the standard sequence-to-sequence model.
A sequence of city coordinates, $\Bx_1,\ldots,\Bx_N$, is fed to a Transformer encoder via linear embedding and positional encoding modules, and its output is fed to a Transformer decoder, which recursively outputs an approximated solution, i.e., the city index sequence, $\pi_1,\ldots,\pi_N$. The attention mechanism in Transformer can evaluate the mutual relationship between all city pairs and, therefore, is very favorable for TSP because the tour needs to be determined while looking at all cities.
\par
Precisely speaking, Bresson et al.~\cite{bresson2021transformer} does {\em not} treat TSP as a sequence-to-sequence task but a {\em set}-to-sequence task. In general applications of Transformers to sequence-to-sequence tasks (e.g., language translation),  
positional encoding is employed to differentiate the different sequences, such as  $\Bx_1, \Bx_2, \Bx_3, \ldots, \Bx_N$ and $\Bx_2, \Bx_1, \Bx_3, \ldots, \Bx_N$. In contrast, Bresson et al.~\cite{bresson2021transformer}, as well as Kool et al.~\cite{kool2018attention}, do not employ positional encoding, as shown in Fig.~\ref{fig:our_model}(b). Without positional encoding, two input sequences $\Bx_1, \Bx_2, \Bx_3, \ldots, \Bx_N$ and $\Bx_2, \Bx_1, \Bx_3, \ldots, \Bx_N$ are treated as exactly the same inputs. Consequently, their solver can give the same solution regardless of the 
 input order of cities. This property is so-called {\em permutation invariant} and seems very reasonable for TSP because the actual input of TSP is a set of cities, and the solution of TSP should not depend on the input order.
\par

However, this paper goes against this traditional approach and experimentally proves that treating TSP as a sequence-to-sequence task is actually better by the introduction of {\em Test-Time Augmentation} (TTA). TTA is a data-augmentation technique. Different from standard data-augmentation techniques for the training phase, TTA is used in the test phase. Fig.~\ref{fig:our_model}(c) illustrates TTA with the sequence-to-sequence model of (a). As noted above, the sequence-to-sequence model of Fig.~\ref{fig:our_model}(a) give different solutions with $\Bx_1, \Bx_2, \Bx_3, \ldots, \Bx_N$ and $\Bx_2, \Bx_1, \Bx_3, \ldots, \Bx_N$. This means if we generate $M$ variants of a single TSP instance by a random permutation process, we have $M$ different solutions from the $M$ variants. Consequently, we will have an accurate solution by choosing the best solution among $M$ solutions.
\par 

This paper proposes a Transformer-based model for solving TSP with the above idea about TTA. Fig.~\ref{fig:our_model}(d) shows our model. In our model, we utilize the fact that a TSP instance can be represented as a $N\times N$ distance matrix of Fig.~\ref{fig:tsp_style}, whose $(i,j)$-th element is the distance $d_{i,j}$ between two cities, $\Bx_i$ and $\Bx_j$. 
If we exchange the first and second cities (i.e., $\Bx_1, \Bx_2, \Bx_3, \ldots, \Bx_N$ to $\Bx_2, \Bx_1, \Bx_3, \ldots, \Bx_N$), the first and second rows and columns are flipped in the matrix. Similarly, by applying a random permutation to the city order, we can generate variants of the distance matrix. As shown in Fig.~\ref{fig:tsp_style}, the distance matrix is treated as a sequence of the column vector, 
$\Bd_1,\ldots, \Bd_N$, where $\Bd_i\in \mathbb{R}^N$. Consequently, by TTA with $M$ random permutations, we have $M$ different column vector sequences, which become $M$ different inputs to the model of Fig.~\ref{fig:our_model}(d) and give $M$ different solutions. 
Since $\Bd_i$ comprises the distances from a certain city to all the $N$ cities, it carries stable information (because $d_{i,j}$ is invariant to the geometric translation and rotation) than $\Bx_i$. Moreover, since the random permutations affect not only columns but also rows of the distance matrix, our column vector representation realizes wider variations, which has a positive effect on TTA.
\par 

We experimentally show that our simple but effective model can achieve a quite low optimality gap of $0.01$\% for the public TSP dataset, TSP50, and $1.07$\% for TSP100. In addition, our result proves that the solution quality monotonically increases by the augmentation size $M$. This property is useful for users because they can set $M$ by considering the trade-off between the solution quality and the computational complexity.\par


The main contributions of this paper are summarized as follows:
\begin{itemize}
\item We propose a machine learning-based model for solving TSP with an elaborated TTA. Specifically, our model first represents the TSP instance as a distance matrix and then generates its multiple variants by random permutations. 
\item Our model outperformed the latest methods on the public TSP instance datasets, TSP50 and TSP100 datasets. 
\end{itemize}
\par 

\section{Related work\label{sec:review}}

\subsection{TSP solver}


Roughly speaking, TSP solvers can be classified into three types: exact solvers, (algorithmic) approximate solvers, and machine learning-based solvers. As an exact solver, we can find a classical Held-Karp algorithm~\cite{Held1962ADP}, which is based on dynamic programming. This algorithm, however, becomes intractable with a large $N$ (say, 40) due to the NP-hardness of TSP. In recent research, Concorde~\cite{applegate2006concorde} has been used as an exact solver. Concorde is based on an algorithm that combines integer programming with cutting planes and branch-and-bound. Note that there are several exact solvers, such as \cite{berg}, which focus only on Euclidean TSP; however, they also become intractable with a large $N$ because TSP is still NP-hard even under the Euclidean metric condition.\par
Approximate solvers are computationally efficient at the expense of solution accuracy.
The classical approximate solver by Christofides et al.~\cite{Christofides1976WorstCaseAO} prepares the minimum spanning tree (with a linear time complexity) and then modifies it to have a tour. Its approximate solution is guaranteed that its tour length is less than $3/2$ of the minimum tour length. Another classical approximate solver is 2-opt~\cite{Johnson1990LocalOA,Lin1965ComputerSO}, where two edges are swapped if the tour becomes shorter. There are solvers which combine Christofides and 2-opt~\cite{Helsgaun2000AnEI,Helsgaun2017AnEO,Lin1973AnEH}.\par

Machine learning-based solvers became alternatives to the above algorithm-based solvers because of the rapid development of deep neural networks. Vinyals et al.~\cite{Vinyals2015} proposed the Pointer Network, which is based on a recurrent neural network (RNN). 
Considering the difficulty of preparing the ground-truth for each training instance, which is NP-hard, Bello et al.~\cite{Bello2016NeuralCO} extended the Pointer Network to be trained in a reinforcement learning framework. Graph neural networks (GCN) are also common networks for solving TSP~\cite{gcn-tsp,kool2018attention}. In these GCN-based models, each node of the input graph corresponds to a city, and each edge connects a pair of (roughly) neighboring cities. Then, GCN aggregates the information of the neighboring cities and finally determines the tour. \par

The majority of the recent machine learning-based TSP solvers use Transformers. An advantage of using the Transformer is its attention mechanism that evaluates the relationships between all $N^2$ pairs of $N$ cities.
Bresson et al.~\cite{bresson2021transformer} proposed a method for predicting tours in the TSP based on the Transformer encoder and decoder model, which has been utilized in language translation. More precisely, as we saw in Fig.~\ref{fig:our_model}(b), their model uses the encoder-decoder model as a {\em set-to-sequence} converter by removing the positional encoding. Jung et al.~\cite{jung2023lightweight} combines a convolutional neural network (CNN) with \cite{bresson2021transformer}. The CNN is responsible for extracting city features in the $k$-nearest neighborhood. Kwon et al.~\cite{kwon2020pomo,kwon2021matrix} also use Transformers to solve ``asymmetric'' TSPs by their bipartite graph representation.
\par


There are two problem representation styles in the above machine learning-based TSP solvers. The first representation style is a set of city coordinates and is employed in most solvers~\cite{Bello2016NeuralCO,bresson2021transformer,jung2023lightweight,Vinyals2015}. This is a straightforward representation because the input of the original TSP is the set of city coordinates. The second style is a $N \times N$ distance matrix that comprises distances for all pairs of $N$ cities. For GCN-based solver, where edges connecting two cities are crucial, employs this representation style. To the authors' knowledge, there is no past attempt that decomposes the distance matrix into a sequence of the $N$ column vector $\Bd_1,\ldots,\Bd_N$.

\subsection{Test-time augmentation (TTA)}
TTA is a technique aimed at improving prediction accuracy by performing data augmentation during inference ~\cite{Kim2020,kimura2021understanding,lyzhov2020greedy,moshkov2020test,shanmugam2021better,wang2019aleatoric}. In general, TTA is a two-step procedure in the testing phase. First, from the input to be tested, $M$ variants are generated by data augmentation techniques. Second, $M$ model predictions for $M$ variants are aggregated by, for example, majority voting or simple max/min operation to determine the final prediction.
Augmentation and aggregation strategies depend on the characteristics of the target dataset and the model. 
Famous applications of TTA are image classification~\cite{Kim2020,lyzhov2020greedy} and image segmentation~\cite{moshkov2020test}. Kimura et al.\cite{kimura2021understanding} have theoretically proved the effectiveness of TTA.
\par

The effectiveness of TTA relies on the lack of invariance in prediction models. In other words, TTA has no effect if the target model is invariant to some specific changes (such as rotation for images and permutation for sequences), all $M$ variants by the changes are identical for the model, and the $M$ solutions are the same. One example of the invariant model is Fig.~\ref{fig:our_model}(b), which depicts an input order-invariant set-to-sequence model, such as Bresson et al.~\cite{bresson2021transformer}.
\par

\section{Test-Time Augmeantion (TTA) for the Traveling Salesperson Problem (TSP)}

\subsection{Overview}
We propose the TTA method for TSP as shown in Fig.~\ref{fig:our_model}(d).
As described in Section~\ref{sec:intro}, the latest Transformer-based solvers~\cite{bresson2021transformer,jung2023lightweight} treat TSP as a {\em set}-to-sequence conversion problem as shown in Fig.~\ref{fig:our_model}(b). In other words, this solver is invariant to the permutation in the input city order and gives the same tour for different inputs, such as  $\Bx_1, \Bx_2, \Bx_3, \ldots, \Bx_N$ and $\Bx_2, \Bx_1, \Bx_3, \ldots, \Bx_N$. In contrast, we treat TSP in a {\em sequence}-to-sequence model. Since we treat the input as a sequence, the model output will be sensitive to the city order of the input. Therefore, we can perform TTA with the model. Specifically, we first generate $M$ variants of a single TSP instance by representing it as a distance matrix and applying random permutations of the input city order. Then, we feed them to the model and have $M$ different solutions (i.e., tours). Finally, among $M$ solutions, we chose the best solution with the minimum tour length.\par

%
Note that our model is based on the model architecture by Bresson et al.~\cite{bresson2021transformer}, while Jung et al.~\cite{jung2023lightweight} is an improved version of \cite{bresson2021transformer}. This is because the model of \cite{bresson2021transformer} is simpler (as shown in Fig.~\ref{fig:our_model}(b)), and therefore, comparative studies will also become simpler. Moreover, another literature~\cite{xiao} shows Bresson et al.~\cite{bresson2021transformer} outperforms \cite{jung2023lightweight} in a certain training condition.\par

\subsection{TTA with distance matrix}
We represent each TSP instance as a $N \times N$ distance matrix and use it as the input of the model. The $(i,j)$-th element of the matrix is the Euclidean distance  $d_{i,j}$  between two cities $\Bx_i$ and $\Bx_j$. As shown in Fig.~\ref{fig:tsp_style}, the distance matrix is treated as a sequence of the $N$ column vectors, $\Bd_1,\dots,\Bd_i,\ldots,\Bd_N$, where $\Bd_i\in \mathbb{R}^N$.\par
Why do we use $\Bd_i$ instead of $\Bx_i$? In fact, if we use $\Bx_i$, we can realize the model more simply, as shown in Fig.~\ref{fig:our_model}(c).
There are several reasons to insist on using $\Bd_i$.
First, the distance $d_{i,j}$ in $\Bd_i$ is a representation invariant to spatial translation and rotation of the city coordinates $\{\Bx_1,\ldots,\Bx_N\}$, whereas the city coordinate $\Bx_i$ is variant to them. The solution of TSP should be invariant to these rigid transformations. Therefore, this invariant property by the distance representation is appropriate for treating TSP in a machine-learning framework. Second, $\Bd_i$ seems like a feature vector representing the positional relationships between the $i$-th city and all $N$ cities. (In fact, a latest model~\cite{jung2023lightweight} introduces a CNN to deal with such relationships among cities.) \par
%
The third and most important reason for using $\Bd_i$ instead of $\Bx_i$ is that $\Bd_i$ shows large variations by random permutation of city indices, and these variations are beneficial for TTA.
The random perturbation affects not only columns but also rows of the distance matrix. Assume that we have a distance matrix $D$ and its column vectors $\Bd_1,\ldots,\Bd_i, \ldots, \Bd_N$, and we have another distance matrix $D'$ by random index permutation and its column vectors $\Bd'_1,\ldots,\Bd'_i, \ldots, \Bd'_N$. Since the permutation affects rows, no column vector in $\{\Bd'_i\}$ is identical to $\Bd_i$. In other words, our TTA by random permutation of city indices is not just a column permutation (like $\Bd_1, \Bd_2, \Bd_3, \ldots, \Bd_N \to \Bd_2, \Bd_1, \Bd_3, \ldots, \Bd_N$) but a more substantial permutation in both of the column and row directions. 

\subsection{Decoding to determine the city order}
The decoder outputs the probability vectors $\Bp_1,\ldots, \Bp_i, \ldots, \Bp_N$. 
Each vector $\Bp_i=(p_{i,1}, \ldots, p_{i,j},\ldots, p_{i,N}) \in \Delta^N$ represents the probability that each city becomes the next city. 
The decoder outputs the probability vectors $\{\Bp_i\}$ in a recursive way from $i=1$ to $N$. Specifically, $\Bp_i$ is given as the output by inputting $\Bp_{i-1}$ to the decoder after the post-processing (shown in Fig.~\ref{fig:our_model}) to mask the cities already visited.  
In this recursive process, we determine the city order in a greedy manner; we choose the city with the highest probability as the $i$-th city to visit, namely, 
\begin{align}
\pi_i = \arg\max_j \{p_{i,1}, \ldots, p_{i,j}, \ldots, p_{i,N}\}.
\end{align}
Through this procedure, our model determines the order of the cities, $\pi_1,\ldots,\pi_N$  from $i=1$ to $N$ in a greedy manner.
\par 

\subsection{Model architecture}
Before feeding these $N$ column vectors into the Transformer encoder, we apply 
traditional positional encoding (PE) with sinusoidal functions~\cite{vaswani2017attention} to individual vectors among various PE strategies~\cite{dufter-etal-2022-position}.
By PE, the Transformer encoder and decoder become input order-variant; namely, 
they can treat $\Bx_1, \Bx_2, \Bx_3, \ldots, \Bx_N$ and $\Bx_2, \Bx_1, \Bx_3, \ldots, \Bx_N$ as different inputs.
\par 

The Transformer encoder has six layers with multi-head attention to extract the relationship between all city pairs, and the Transformer decoder has two layers. The dimension of the latent variables (i.e., encoder outputs) is $512$. Note that these model parameters are determined by following the latest and most famous model~\cite{bresson2021transformer}.
\par 

\subsection{Optimization of the model}
Our model is trained not by supervised learning but by deep reinforcement learning, like the past attempts~\cite{bresson2021transformer,jung2023lightweight}.
This is because it is difficult to obtain a ground-truth due to its computational complexity. In deep reinforcement learning, the city order $\pi_1, \ldots, \pi_N$ is learned by using the tour length as the reward. We use the traditional REINFORCE algorithm~\cite{williams1992simple} to train our model.
\par 
\section{Experiment Setup}
\subsection{Datasets and comparative models}
For a fair comparison with the latest and most famous TSP solver by Bresson et al.~\cite{bresson2021transformer}, we follow their data preparation procedure. 
Each TSP instance is a set of $N=50$ or $100$ cities on a two-dimensional square plane $[0,1]^2$. Therefore, each city coordinate $\Bx_i$ is represented by a two-dimensional vector. We consider Euclidean TSP, where Euclidean distance is used to evaluate the inter-city distance $d_{i,j}=\|\Bx_i-\Bx_j\|$ and therefore the tour length.
\par 
A training set for $N=50$ was comprised of 100,000 instances, and each instance was comprised of 50 points randomly generated in $[0,1]^2$. A training set for $N=100$ was prepared in the same manner. Note that we do not know the ground-truth, i.e., the minimum length tour, of each instance. We, therefore, needed to use a reinforcement learning framework for training the model.
\par 
For testing, we used TSP50 and TSP100~\cite{bresson2021transformer} for $N=50$ and $100$, respectively, which are publicly available\footnote{\tt https://github.com/xbresson/TSP\_Transformer}. Each of them contains 10,000 instances. 
For each test instance, a ground-truth is provided by using Concorde~\cite{applegate2006concorde}, which is a well-known algorithm-based TSP solver.
\par  

\subsection{Implementation details}
We followed the past attempts~\cite{bresson2021transformer,jung2023lightweight} for setting various hyperparameters, such as the model architecture (the number of layers in the encoder and decoder, the vector dimensions, etc.), learning rate, optimizer, batch size, and the number of the training epochs.
\par 

We also follow the past attempts ~\cite{bresson2021transformer,jung2023lightweight} 
for training our model in a deep reinforcement learning framework. 
Specifically, we used the REINFORCE algorithm~\cite{williams1992simple}. 
Since reinforcement learning requires extensive computation time, we terminated the training process with 100 epochs by following~\cite{jung2023lightweight}.
\par 

For our model, we set its default augmented size $M$ at $2500$. We also conducted an experiment to see the effect of $M$; in the experiment, we changed $M$ from 2 to 1024. As we will see later, the performance of our model is monotonically improved according to $M$.
\par 

\subsection{Performance metrics}
By following the previous attenpts~\cite{bresson2021transformer,gcn-tsp,jung2023lightweight,kool2018attention}, the performance on the test data is evaluated by the optimality gap (\%) and the average tour length:
\begin{align}
&\mathrm{Optimality\ gap}= 
   \frac{1}{K} \sum_{k=1}^{K}  \left( \frac{\hat{l}_{k}^{\mathrm{TSP}}}{l_{k}^{\mathrm{TSP}}}  - 1\right), \\  
&\mathrm{Average\ tour\ length} = \frac{1}{K}\sum_{k=1}^{K}\hat{l}_{k}^{\mathrm{TSP}},
\end{align}
where $K$ is the number of instances, 
$\hat{l}_{k}^{\mathrm{TSP}}$ is the predicted tour length for the $k$-th instance, and $l_{k}^{\mathrm{TSP}}$ is the shortest tour length provided by Concorde~\cite{applegate2006concorde}. 
\par

\subsection{Comparative models}
We compared the following two latest solvers, Bresson et al.~\cite{bresson2021transformer} and Jung et al.~\cite{jung2023lightweight}. As noted before, both solvers are based on the Transformer encoder and decoder, and the latter introduces a CNN into the former framework. In other words, the latter is an improved version of the former. We employ them as comparative methods because they are the latest TSP solvers whose input is a set of city coordinates $\{\Bx_1,\ldots,\Bx_N\}$.
\par 

For a fair comparison with these latest solvers, which do not employ TTA, 
we employ the beam search in the decoding process. By the beam search, we can obtain $B$ tour candidates, where $B$ is called ``beam width''. Each tour candidate has a different tour length,  and we choose the one with the minimum tour length as our solution. The latest solvers \cite{bresson2021transformer,jung2023lightweight} have already introduced this beam search to improve their solution. Since we set the default augmented size $M=2500$, 
we also set the default beam size $B=2500$. In the later experiment, we also evaluated the performance of these solvers with $B=1$, which resulted in the greedy search for a minimum-length tour like ours. 
\par 

\begin{center}
\begin{table}[t]
\caption{Average tour length (obj.) and optimality gap (Gap.).
G and BS stand for the greedy search and the beam search in the decoding process, respectively.}
\centering
\begin{tabular*}{0.9\linewidth}{@{\extracolsep{\fill}} lc c cc cc}
    \thickhline
  \multirow{2}{*}{Method}   &  Decoding  & \multicolumn{2}{c}{ TSP50 }  & \multicolumn{2}{c}{ TSP100 }   \\ 
    &process  & Obj. & Gap.  & Obj. & Gap.  \\ \hline \hline  
  Concorde\cite{applegate2006concorde} 
  &(Exact Solver) & 5.690  & 0.00\% & 7.765 &  0.00\% \\ \hline
  
  \multirow{2}{*}{Bresson et al. \cite{bresson2021transformer}}& \typeG & 5.754  & 1.12\% & 8.005 &  3.09\% \\
& \typeB & 5.698  &  0.14\% &  7.862 & 1.25\% \\ \hline

\multirow{2}{*}{Jung et al.  \cite{jung2023lightweight}} & \typeG & 5.745  &   0.97\%   & 7.985 & 2.83\% \\
 &\typeB &  5.695   &  0.10\%  & 7.851 &  1.11\% \\ \hline 

Our model  &\typeG & \textbf{5.690}          &  \textbf{0.01\%} & \textbf{7.848} & \textbf{1.07\%} \\ \hline
~~~  w/o Distance Matrix &\typeG & 5.731   &  0.72\% &7.968 & 2.61\% \\ 
\hline
\multirow{2}{*}{~~~ w/o TTA}  & \typeG& 5.837        & 2.58\%  & 8.283 &   6.68\% \\ 
  &\typeB & 5.720          & 0.53\%  & 8.039 & 3.53\%   \\ 
      \thickhline
\end{tabular*}
\label{table:legth_gap}
\end{table}
\end{center}

\section{Experimental results}

\begin{figure}[t]
    \centering
    \vspace{5mm}
    \includegraphics[width=0.92\linewidth]{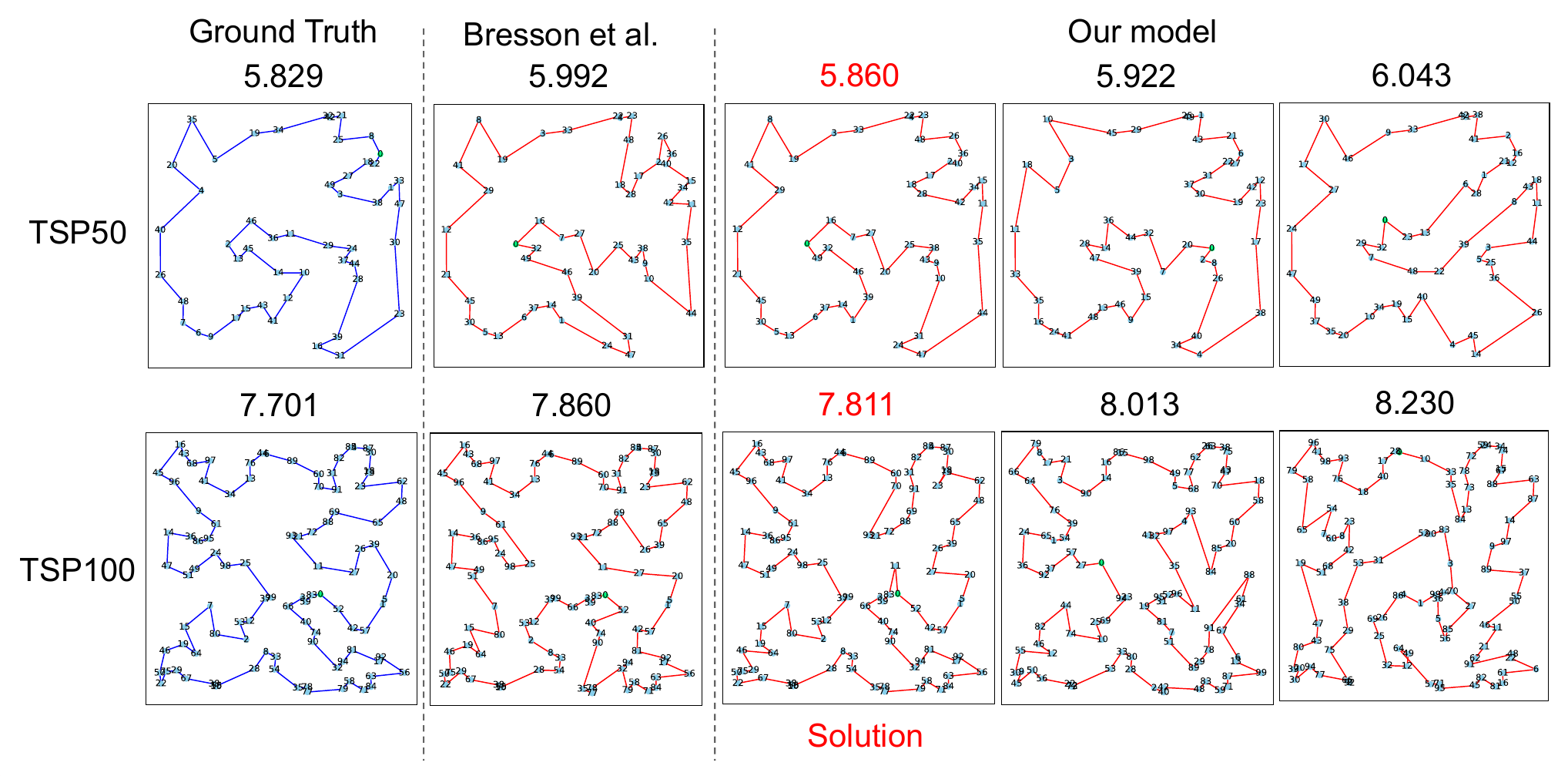}
    \caption{Solution examples in TSP50 and TSP100. The number indicates the tour length. Our model assumes $M=3$.}
    \label{fig:networks}
\end{figure}

\subsection{Comparative experiments with the conventional TSP solvers}
Table~\ref{table:legth_gap} shows the average tour length (``Obj.'') and optimality gap (``Gap.'') on TSP50 and TSP100. ``G'' and ``BS'' stand for the greedy search and the beam search in the decoding process of the comparative models. The results show that our model outperforms two comparative models in both datasets and in both metrics. In particular, in TSP50, our method archives almost the same tour length as the exact solver.
\par 

As an ablation study, we evaluated our model without using the distance matrix. More specifically, we used the city coordinates $\{\Bx_i\}$ instead of the column vectors $\{\Bd_i\}$. We used positional encoding for $\Bx_i$, and therefore, we still performed TTA in this ablation setup. (Note that this setup can be seen as an extended version of Bresson et al.~\cite{bresson2021transformer} by introducing the same positional encoding and then TTA.) Our model shows much better performance than this ablation setup (``Our model w/o Distance Matrix''). Consequently, using the distance matrix as an ordered sequence of the column vectors $\Bd_1,\ldots,\Bd_N$ is proved to be useful in the TTA framework.
\par
As another ablation study, we evaluated our model without TTA. The result in Table~\ref{table:legth_gap} shows that the performance of our model degrades drastically by the omission of TTA. In other words, this ablation study emphasizes how TTA is effective in our model. (Note that the performance drop by the ablation of TTA is far more serious than the drop by the above ablation of the distance matrix.)
As described before, random city index permutation affects the distance matrix in both the column and row directions. This large effect of permutation helps TTA find a better solution. We also evaluated our model without TTA but with beam search; its performance is still much lower than our model with TTA. This difference also proves the effectiveness of TTA with the distance matrix. 
\par 

Fig.~\ref{fig:networks} illustrates the predicted tour for a specific instance of TSP50 and TSP100. Our model predicted different tours by TTA. Compared with Bresson et al.~\cite{bresson2021transformer}, the tour shown in the center of Fig.~\ref{fig:networks} by our model resembles the ground-truth by Concorde and thus gives a shorter (i.e., better) tour.
\par 

\subsection{Effect of augmentation size}

\begin{figure}[t]
    \centering
    \includegraphics[width=\linewidth]{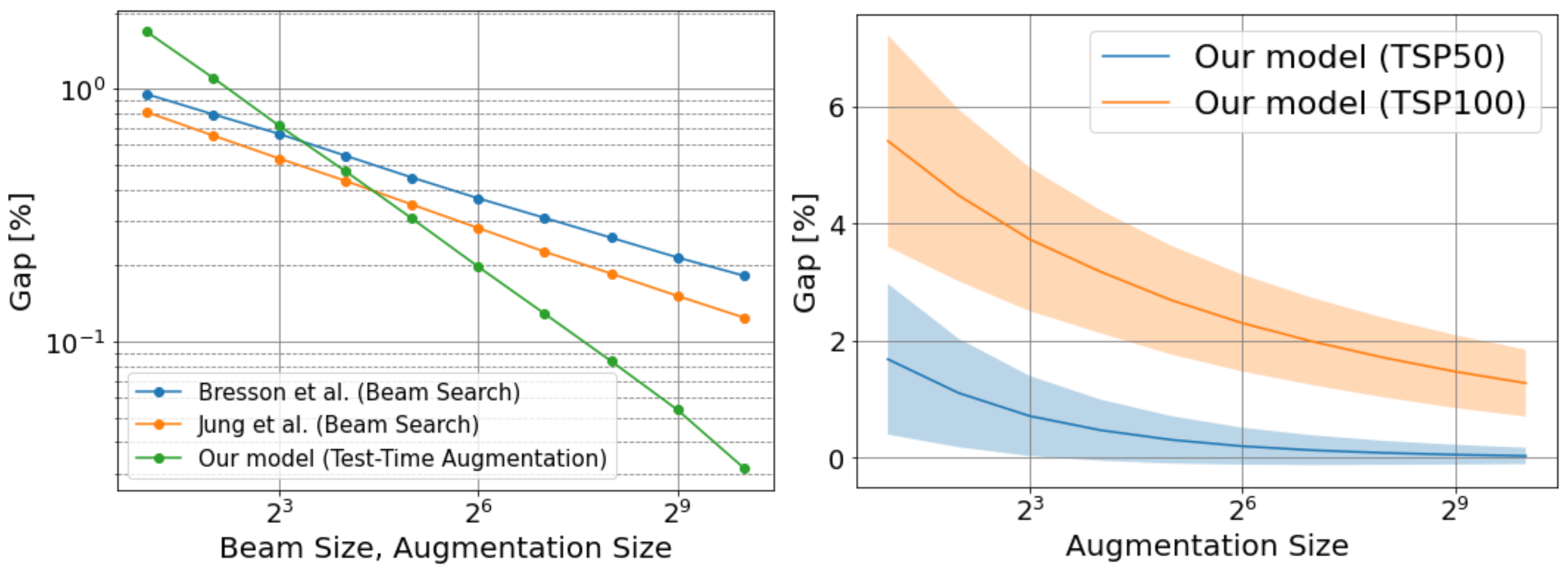}
    \caption{Left: Effect of the augmentation size $M$ on the performance of our model on TSP50. The performance of the latest models~\cite{bresson2021transformer,jung2023lightweight} under different beam width $B$ is also plotted. Right: Performance variations in all instances in TSP50 or TSP100. Shaded regions indicate the standard deviation intervals. Note that the vertical axis of the left plot is logarithmic, whereas that of the right plot is not.}
    \label{fig:bs_vs_tta_avg}
\end{figure}

Fig.\ref{fig:bs_vs_tta_avg}(Left) shows how the augmentation size $M$ affects the performance of our model on TSP50. By increasing $M$, the performance (Gap) of our model is monotonically improved. This result also proves the effectiveness of TTA in improving the accuracy. The computational cost increases with $M$. Therefore, this result shows that we have a simple trade-off between the computational cost and accuracy. This property is important for the practical use of our model. (In other words, if this curve is non-monotonic, it is difficult to determine the best compromise between computations and accuracy.)
\par 

Fig.\ref{fig:bs_vs_tta_avg}(Left) also shows the performance
of the latest models~\cite{bresson2021transformer,jung2023lightweight} under different beam width $B$. Like our model, these models also show monotonic performance improvements according to $B$. However, the degree of their improvements is not as large as ours. Note that, in this graph, all three models show a near-linear relationship between the gap (i.e., accuracy) and the augmented size $M$ (or the beam size $B$) in this logarithmic plot. This means that the gap $\propto$ $e^{-\alpha M}$ (or $e^{-\beta B}$), where $\alpha$ (or $\beta$) is a positive constant representing the slope of the graph. In addition, since the graph shows $\alpha > \beta$, the increase of $M$ is more effective than that of $B$, although it is not meaningful to compare $M$ and $B$ directly.
\par 


Fig.\ref{fig:bs_vs_tta_avg}(right) shows the performance variations in all instances of TSP50 and TSP100 by our model. As the augmentation size $M$ increases, the variation decreases; this means that more accurate paths are obtained in most instances. Note that the curve of TSP100 does not show performance saturation. Larger $M$ will give better tours even for more difficult tasks with more cities, $N$.
\par 

\subsection{Instance-level performance evaluation}

\begin{figure}[t]
    \centering
    \includegraphics[width=0.7\linewidth]{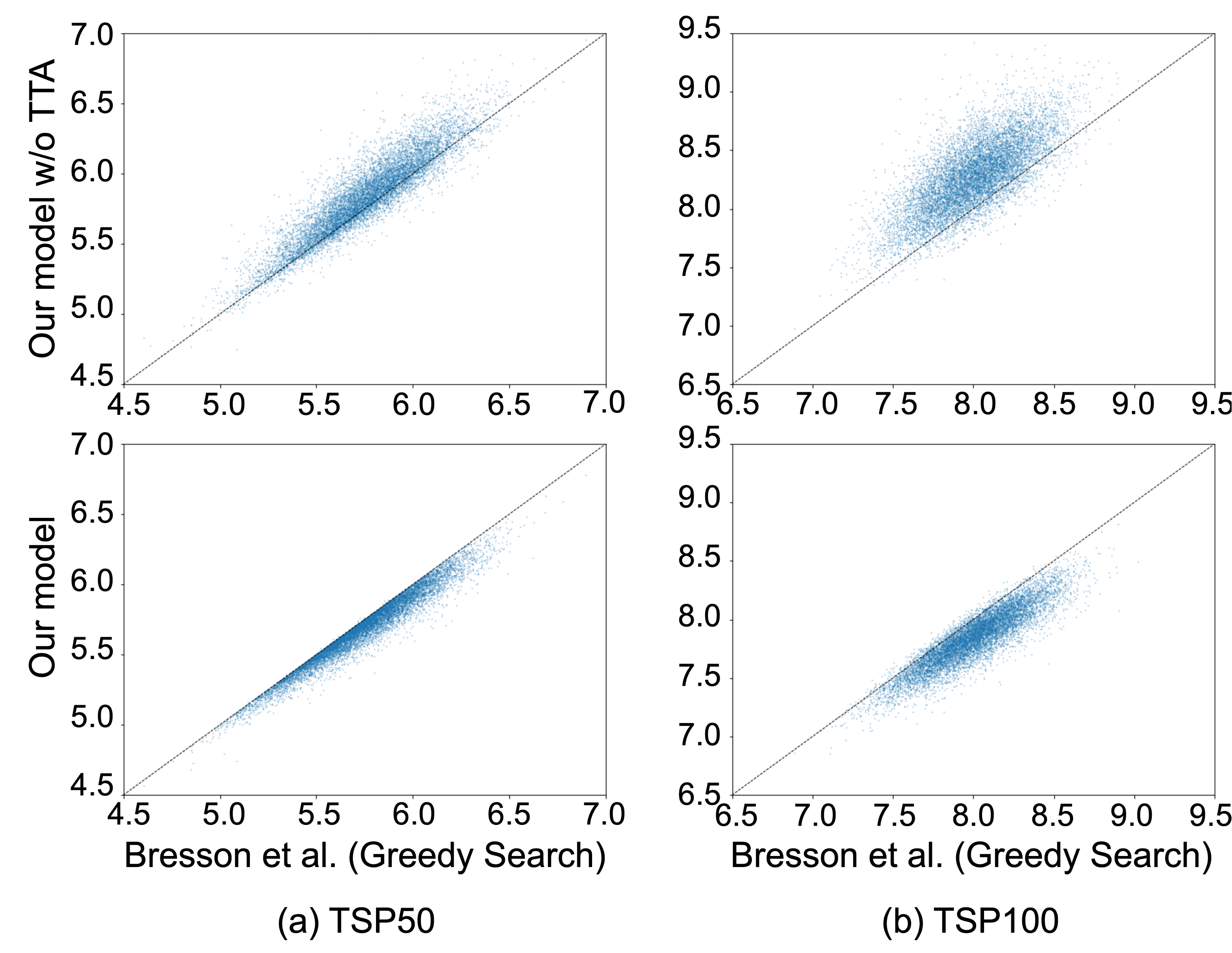}
    \caption{Instance-level comparison with Bresson et al.~\cite{bresson2021transformer}. Here, the tour length is used as the performance metric. In the upper plots, the vertical axes correspond to the performance of our model without TTA, whereas the horizontal axes correspond to Bresson et al. In the lower plots, the vertical axes show the performance of our model (with TTA). By comparing the upper and lower plots, we can also observe the effect of TTA on our model.} 
    \label{fig:inst_level}
\end{figure}

Figs.~\ref{fig:inst_level}(a) and (b) illustrate the instance-level performance on TSP50 and TSP100, respectively. In these plots, each dot corresponds to a TSP instance. The horizontal and vertical axes are the tour length by Bresson et al. and our model, respectively. About our model, the upper plots show the case without TTA and the lower with TTA. The upper plot of (a) shows that Bresson et al. outperformed our model without TTA in most instances; however, the lower plot shows that our model with TTA outperformed Bresson et al. in almost all instances. The same observation can be made in (b). Consequently, from these plots, we can confirm that TTA is beneficial for our model, and our model with TTA is better than Bresson et al. regardless of the distributions of cities.\par

\section{Conclusion, limitation, and future work}
In this study, we introduce test-time data augmentation (TTA) to a machine learning-based traveling salesperson problem (TSP) solver. For the introduction, we reformulated TSP from a set-to-sequence task to a sequence-to-sequence task by (re)employing positional encoding. Moreover, we represent a TSP instance by a distance matrix; more specifically, we represent a TSP instance as a sequence of the column vectors of the matrix. Then, we apply random permutation of city order, which causes variants of the original distance matrix in both the column and row directions. By choosing the best solution among the solutions for the individual variants, we finally have an accurate solution for the instance. Through various experiments, we confirmed the effectiveness of TTA. Moreover, we also confirmed that our model outperforms the latest solvers~\cite{bresson2021transformer,jung2023lightweight}.
\par

The limitation of the current model is that it assumes a fixed city number by following the tradition of machine learning-based TSP solvers. Research on approximation methods using machine learning for the TSP often assumes variable numbers of cities~\cite{fu2021generalize,joshi2022learning}. However, the current model relies on the fixed size of the distance matrix because we assume the dimension of the column vector $\Bd_i$ is fixed. (Note that the basic architecture of the Transformer encoder and decoder can accept variable $N$.) Therefore, to deal with the arbitrary number of cities, $N$, we need to employ some module (such as a Transformer) that can convert $\Bd_i$ (with an arbitrary dimension $N$) into a fixed dimensional vector. 
\par

There are several other future works. First, one may introduce a learning scheme to find the optimal TTA strategies. In this paper, we use TTA based on uniform random permutations. In contrast, we can find trainable TTA strategies, especially in recent image processing research fields~\cite{lyzhov2020greedy,shanmugam2021better}. Second, one may develop a more computationally 
efficient TTA. Through the experimental results in this paper, we found that the accuracy improves along with the augmentation size $M$. Therefore, we want to increase $M$ as many as possible, but the computational cost increases linearly with $M$. To relax this trade-off, we can introduce some algorithmic methods, as well as the trainable TTA strategies mentioned above.
\par

\clearpage
%


\bibliographystyle{splncs04}
\bibliography{mybib}

\end{document}